\documentclass{article}
\usepackage[margin=1in]{geometry}
\usepackage[T1]{fontenc}
\usepackage{graphicx} 
\usepackage{amsthm, amsmath, amssymb, bm, natbib}
\usepackage{xcolor}
\usepackage{multirow, makecell}
\usepackage{hyperref, cleveref}
\usepackage{booktabs}
\usepackage{booktabs}  
\usepackage{makecell}  
\usepackage{multirow}  

\theoremstyle{definition}
\newtheorem{definition}{Definition}
\theoremstyle{theorem}
\newtheorem{lemma}{Lemma}
\newtheorem{theorem}{Theorem}
\renewcommand{\thefootnote}{\fnsymbol{footnote}} 

\title{Branching Strategies Based on Subgraph GNNs: A Study on Theoretical Promise versus Practical Reality}
\author{
  Junru Zhou\textsuperscript{1,*}, 
  Yicheng Wang\textsuperscript{2,*}, 
  Pan Li\textsuperscript{3}}
\date{Technical Report | Summer 2024} 

\begin{document}
\maketitle
{
  \centering
  \textsuperscript{1}Peking University \quad 
  \textsuperscript{2}HKUST \quad 
  \textsuperscript{3}Georgia Tech \\
}

\begin{abstract}
Graph Neural Networks (GNNs) have emerged as a promising approach for ``learning to branch'' in Mixed-Integer Linear Programming (MILP). While standard Message-Passing GNNs (MPNNs) are efficient, they theoretically lack the expressive power to fully represent MILP structures. Conversely, higher-order GNNs (like 2-FGNNs) are expressive but computationally prohibitive. In this work, we investigate Subgraph GNNs as a theoretical middle ground. Crucially, while previous work \citep{chen2025rethinkingcapacitygraphneural} demonstrated that GNNs with 3-WL expressive power can approximate Strong Branching, we prove a sharper result: node-anchored Subgraph GNNs whose expressive power is strictly lower than 3-WL \citep{zhang2023completeexpressivenesshierarchysubgraph} are sufficient to approximate Strong Branching scores. However, our extensive empirical evaluation on four benchmark datasets reveals a stark contrast between theory and practice. While node-anchored Subgraph GNNs theoretically offer superior branching decisions, their $O(n)$ factor complexity overhead results in significant memory bottlenecks and slower solving times compared to MPNNs and heuristics. Our results indicate that for MILP branching, the computational cost of expressive GNNs currently outweighs their gains in decision quality, suggesting that future research must focus on efficiency-preserving expressivity.
\end{abstract}

\footnotetext[1]{These authors contributed equally to this work. Work performed during a Summer 2024 internship. Emails: \texttt{zml72062@stu.pku.edu.cn}, \texttt{yw112@illinois.edu}, \texttt{panli@gatech.edu}}

\setcounter{footnote}{0} 
\renewcommand{\thefootnote}{\arabic{footnote}} 

\section{Introduction}
A vast number of real-world problems in fields such as management, retail, and manufacturing can be formulated as combinatorial optimization problems~\citep{paschos2014applications}. Combinatorial optimization seeks to optimize objective functions over discrete domains, where brute-force enumeration is intractable. Many such problems, including Set Covering and Maximum Independent Set, are known to be NP-hard~\citep{karp2009reducibility}, admitting no polynomial-time deterministic algorithms unless $\text{P}=\text{NP}$.

Despite this theoretical intractability, a plethora of practical algorithms exists to tackle these problems effectively. Although they possess exponential worst-case time complexity, algorithms like \textbf{Branch-and-Bound (B\&B)} are often able to efficiently find optimal solutions for small- or medium-scale instances~\citep{wolsey2020integer}. Instead of blindly searching the entire solution space, B\&B uses heuristic functions to estimate the best possible objective value for the current branch, pruning it if it cannot improve upon the incumbent solution. The efficiency of B\&B relies heavily on the quality of these heuristics.

Many combinatorial optimization problems are formally stated as \textbf{Mixed-Integer Linear Programming (MILP)} problems. B\&B algorithms are especially suitable for MILPs, as one can estimate bounds by solving the \emph{relaxation}, a Linear Programming (LP) problem formed by removing integral constraints~\citep{land2009automatic}. The efficiency of B\&B on MILPs hinges on two decision-making stages~\citep{lodi2017}: \textbf{node selection} (choosing the next sub-problem to solve) and \textbf{variable selection} or \textbf{branching} (choosing a variable to split the search space). Traditionally, these decisions rely on human-designed heuristics, such as \emph{reliability branching} implemented in solvers like SCIP~\citep{a2e5d9e8f52a47c183e8ad12e65499e5} (evaluated in benchmarks such as \citep{koch2011}). While successful, these heuristics are inherently \emph{data-agnostic}, failing to exploit specific statistical patterns of the problem instances at hand.

More recently, the paradigm of \textbf{learning to branch}~\citep{gasse2019exactcombinatorialoptimizationgraph,gupta2020hybridmodelslearningbranch,hansknecht2018cutsprimalheuristicslearning,Khalil_LeBodic_Song_Nemhauser_Dilkina_2016,nair2021solvingmixedintegerprograms} has emerged to address this limitation. These strategies employ machine learning models to predict the optimal branching variable given the solver's history. Among these models, Graph Neural Networks (GNNs) are of particular interest because graphs provide a natural, permutation-invariant representation for MILP problems~\citep{gasse2019exactcombinatorialoptimizationgraph}.

However, a critical question remains regarding the alignment between GNN architectures and MILP algorithms. \citet{chen2023representinglinearprogramsgraph} and \citet{chen2023representingmixedintegerlinearprograms} first studied this from the perspective of \emph{expressive power}. They demonstrated that standard message-passing GNNs (MPNNs)~\citep{xu2019powerfulgraphneuralnetworks} are expressive enough to solve arbitrary LPs (aligning with interior-point methods~\citep{qian2023exploringpowergraphneural}), but are theoretically insufficient for general MILPs. This limitation spurred interest in higher-order architectures. \citet{chen2025rethinkingcapacitygraphneural} recently proposed that \textbf{Second-Order Folklore GNNs (2-FGNNs)}~\cite{maron2019provably} are expressive enough to approximate \textbf{Strong Branching}, an empirically superior but computationally expensive heuristic.

While \citet{chen2025rethinkingcapacitygraphneural} established 2-FGNNs as a theoretical candidate, their practical viability is questionable. 2-FGNNs incur a high time and space complexity, making them difficult to deploy in iterative solvers. This raises the need for a ``middle ground'' architecture.

In this work, we investigate (node-anchored) Subgraph GNNs~\citep{li2020distance,zhang2021nested,you2021identityawaregraphneuralnetworks,zhang2023completeexpressivenesshierarchysubgraph,bevilacqua2022equivariant} as this theoretical middle ground. We improve upon the theoretical result of \citet{chen2025rethinkingcapacitygraphneural} by proving a sharper bound: while 2-FGNNs rely on the expressive power of the 3-WL test, node-anchored Subgraph GNNs, whose expressive power is strictly lower than 3-WL, are sufficient to approximate Strong Branching and empirically more efficient in computation. 

However, theoretical sufficiency does not always imply practical utility. We conduct extensive experiments on four MILP datasets (Set Covering, Combinatorial Auction, Capacitated Facility Location, and Maximum Independent Set) to verify if Subgraph GNNs can overcome the computational barriers of 2-FGNNs. Our results reveal a stark contrast between theory and practice:
\begin{itemize}
    \item Theoretical Advancement: We prove that Subgraph GNNs are expressive enough to theoretically approximate the strong branching score, identifying them as a more efficient theoretical alternative to 2-FGNNs.
 \item \textbf{Complexity Analysis:} We detail the complexity hierarchy, demonstrating that while Subgraph GNNs ($O(nEL)$) scale more efficiently than 2-FGNNs ($O(n^2(n+m)L)$) on the sparse graph structures typical of MILP instances, they still impose a significant overhead compared to standard MPNNs ($O(EL)$). Here, $n$ and $m$ denote the total numbers of variables and constraints, respectively. And, $E$ represents the number of non-zero entries in the constraint matrix.
    \item Empirical Reality Check: We report that despite their theoretical expressivity and improved efficiency compared to 2-FGNNs, Subgraph GNNs still struggle in practice. They frequently encounter Out-Of-Memory (OOM) errors on denser instances and incur slower solving times without consistently reducing search tree sizes, suggesting that the computational overhead of expressive GNNs currently outweighs their decision-making benefits in MILP branching.
\end{itemize}

\section{Preliminaries}

\subsection{Mixed-integer linear programming}

We consider \textbf{mixed-integer linear programming (MILP)} problems of the following \textbf{standard form}.

\begin{definition}[Mixed-integer linear programming, MILP]
    A \textbf{mixed-integer linear programming} problem is one that takes the form
    \begin{align}
        \begin{array}{rl}
        \min_{\bm{x}\in\mathbb{R}^n} \ & \bm{c}^T \bm{x} \\
        &\\
        \mbox{such that} \ & \bm{A} \bm{x} \leqslant \bm{b},\\
        & \bm{l} \leqslant \bm{x} \leqslant \bm{u} ,\\
        & x_1,\ldots, x_r\in\mathbb{Z},
        \end{array}\label{eq:def-milp}
    \end{align}
    where we assume $\bm{x}\in\mathbb{R}^n$ is a vector of $n$ \textbf{variables}, $\bm{A}\in\mathbb{R}^{m\times n}$ together with $\bm{b}\in\mathbb{R}^m$ defines $m$ inequality \textbf{constraints}, and $\bm{l}\in(\{-\infty\}\cup\mathbb{R})^n$, $\bm{u}\in(\mathbb{R}\cup\{+\infty\})^n$ define the \textbf{lower bound} and \textbf{upper bound} of each variable. Finally, the first $r$ variables $x_1,\ldots, x_r$ ($0\leqslant r\leqslant n$) are restricted to be integral.
\end{definition}

If the MILP problem \eqref{eq:def-milp} is infeasible, its optimal objective value is defined as $+\infty$. If the objective of \eqref{eq:def-milp} can be arbitraily lowered, its optimal objective value is defined as $-\infty$.

\subsection{A Brief Review of Branching}

The standard exact method for solving MILPs is the \textbf{Branch-and-Bound (B\&B)} algorithm. It relies on solving the \textbf{LP relaxation} of \eqref{eq:def-milp}, denoted as $\mathcal{P}_{\text{LP}}$, which is obtained by removing the integrality constraints ($x_j \in \mathbb{Z}$).

\begin{definition}[LP relaxation]
    Given an MILP problem \eqref{eq:def-milp}, its \textbf{LP relaxation} is defined as 
    \begin{align}
        \begin{array}{rl}
        \min_{\bm{x}\in\mathbb{R}^n} \ & \bm{c}^T \bm{x} \\
        &\\
        \mbox{such that} \ & \bm{A} \bm{x} \leqslant \bm{b},\\
        & \bm{l} \leqslant \bm{x} \leqslant \bm{u},
        \end{array}\label{eq:lp-relax}
    \end{align}
    which is a linear programming (LP) problem identical to \eqref{eq:def-milp} but with all integrality constraints removed.
\end{definition}

The branch-and-bound algorithm for solving \eqref{eq:def-milp} goes as follows. Instead of directly tackling \eqref{eq:def-milp}, the algorithm solves its LP relaxation \eqref{eq:lp-relax}. Let us assume that the solution\footnote{We assume that \eqref{eq:lp-relax} is feasible, as otherwise the target MILP problem \eqref{eq:def-milp} is definitely infeasible.} to the LP relaxation is $\bm{x}^*$. An obvious truth is that

\begin{lemma}\label{lemma:relax}
    The optimal objective value of \eqref{eq:lp-relax} is never greater than that of \eqref{eq:def-milp}.
\end{lemma}

Now one of the following two cases must occur.

\begin{itemize}
    \item The first $r$ components $x_1^*, \ldots, x_r^*$ of $\bm{x}^*$ are all integral. If that is the case, we actually find the solution to the target MILP problem \eqref{eq:def-milp}---it's simply $\bm{x}^*$.
    \item Not all of $x_1^*, \ldots, x_r^*$ are integral. In this case, the problem \eqref{eq:def-milp} is not yet solved, and we proceed by executing a \textbf{branching} step.
\end{itemize}

\begin{definition}[Branching]
    Assume that $x_{i_1}, \ldots, x_{i_t}$ are the $t$ variables subject to integral constraints in \eqref{eq:def-milp}, yet get non-integral values $x_{i_1}^*,\ldots, x_{i_t}^*$ from the solution $\bm{x}^*$ of the LP relaxation \eqref{eq:lp-relax}. A \textbf{branching} step on LP relaxation \eqref{eq:lp-relax} does the following: first, select a variable $x_{i_j}$ ($j\in\{1,\ldots, t\}$); then solve two LP problems, one of them being
    \begin{align}
        \begin{array}{rl}
        \min_{\bm{x}\in\mathbb{R}^n} \ & \bm{c}^T \bm{x} \\
        &\\
        \mbox{such that} \ & \bm{A} \bm{x} \leqslant \bm{b},\\
        & \bm{l} \leqslant \bm{x} \leqslant \bm{u},\\
        & x_{i_j}\geqslant \lceil x_{i_j}^*\rceil,
        \end{array}\label{eq:up-round}
    \end{align}
    the other being
    \begin{align}
        \begin{array}{rl}
        \min_{\bm{x}\in\mathbb{R}^n} \ & \bm{c}^T \bm{x} \\
        &\\
        \mbox{such that} \ & \bm{A} \bm{x} \leqslant \bm{b},\\
        & \bm{l} \leqslant \bm{x} \leqslant \bm{u},\\
        & x_{i_j}\leqslant \lfloor x_{i_j}^*\rfloor.
        \end{array}\label{eq:below-round}
    \end{align}
\end{definition}

Similar to Lemma \ref{lemma:relax}, we have for LP problems \eqref{eq:up-round} and \eqref{eq:below-round} the following

\begin{lemma}
    The optimal objective value of at least one of \eqref{eq:up-round} and \eqref{eq:below-round} is never greater than that of \eqref{eq:def-milp}.
\end{lemma}

\begin{lemma}
    The optimal objective value of \eqref{eq:lp-relax} is never greater than that of \eqref{eq:up-round} or \eqref{eq:below-round}.
\end{lemma}

Let us continue our discussion of the branch-and-bound algorithm. For the LP problem \eqref{eq:up-round} or \eqref{eq:below-round}, there are still two possible cases.

\begin{itemize}
    \item The solution $\tilde{\bm{x}}^*$ of \eqref{eq:up-round} or \eqref{eq:below-round} has its first $r$ components $\tilde{x}^*_1,\ldots, \tilde{x}^*_r$ all integral. In this case, $\bm{c}^T\tilde{\bm{x}}^*$ gives an \textbf{upper bound} of the optimal objective value of the target MILP problem \eqref{eq:def-milp}.\footnote{We remark that for the case in which \eqref{eq:up-round} or \eqref{eq:below-round} is infeasible, similar discussion also holds, since $+\infty$ would be a trivial upper bound of the optimal objective value of \eqref{eq:def-milp}.} 
    \item Some of the components $\tilde{x}^*_1,\ldots, \tilde{x}^*_r$ are non-integral in the solution $\tilde{\bm{x}}^*$ of \eqref{eq:up-round} or \eqref{eq:below-round}. When this is the case, a further branching step would be executed on \eqref{eq:up-round} or \eqref{eq:below-round}.
\end{itemize}

As we proceed with more branching steps, a series of LP problems are produced. Those LP problems are structured into a \textbf{tree}. For a leaf LP problem, we call it \textbf{closed}, if any of the following conditions holds: (a) it is infeasible, (b) its solution satisfies all integrality constraints as per \eqref{eq:def-milp}, or (c) its optimal objective value is greater than the lowest objective value of all feasible solutions of \eqref{eq:def-milp} known so far. Otherwise, the leaf LP problem is called \textbf{open}. The minimum optimal objective values of closed leaf LPs constitutes an upper bound of the optimal objective value of \eqref{eq:def-milp}, while the minimum optimal objective value of open leaf LPs constitutes a lower bound of the optimal objective value of \eqref{eq:def-milp}. 

Once a leaf LP is considered closed, it either terminates at a feasible solution of the original MILP problem \eqref{eq:def-milp}, or is deemed impossible to produce in its subtree a solution better than the best feasible one obtained so far. Therefore, the algorithm no longer searches its subtree further, or the leaf is \textbf{pruned}. As long as the searching procedure has not finished, the algorithm would find an open leaf LP (via a specific node selection strategy, which is out of the scope of our discussion) and performs another branching step on it, expanding it into a subtree.

\subsection{Strong branching}

A \textbf{branching strategy} is an algorithmic way to select the non-integral variable $x_{i_j}$ in a branching step. A good branching strategy often greatly reduces the search tree size of the B\&B algorithm.

\textbf{Strong branching} (see, for example, \citep{dey2021theoreticalcomputationalanalysisstrongbranching}) is a heuristic branching strategy defined as follows.

\begin{definition}[Strong branching]
    Let $x_{i_1},\ldots, x_{i_t}$ be the variables subject to integral constraints in \eqref{eq:def-milp} yet getting non-integral values $x_{i_1}^*,\ldots, x_{i_t}^*$ from the optimal solution $\bm{x}^*$ of the LP relaxation \eqref{eq:lp-relax}. The strong branching strategy selects the branching variable $x_{i_j}$ ($j\in\{1,\ldots, t\}$) according to the following rule. First, compute a \textbf{score} for each $x_{i_j}$, $j=1,\ldots, t$, 
    \begin{align}
        \text{score}(j)=f(\Delta_j^+,\Delta_j^-),\label{eq:sb-score}
    \end{align}
    where $\Delta_j^+$ means the difference between the optimal objective values of LP problems \eqref{eq:up-round} and \eqref{eq:lp-relax}, $\Delta_j^-$ means the difference between the optimal objective values of LP problems \eqref{eq:below-round} and \eqref{eq:lp-relax}, and $f$ is a hand-designed symmetric function continuous with respect to both its arguments. Two common choices of $f$ are
    \begin{align}
        f(\Delta_j^+,\Delta_j^-)=(1-\mu)\cdot\min\{\Delta_j^+,\Delta_j^-\}+\mu\cdot\max\{\Delta_j^+,\Delta_j^-\},
    \end{align}
    where $\mu\in[0,1]$ is a constant, and
    \begin{align}
        f(\Delta_j^+,\Delta_j^-)=\max\{\Delta_j^+,\epsilon\}\cdot \max\{\Delta_j^-,\epsilon\},
    \end{align}
    where $\epsilon$ is a small constant (e.g., $10^{-6}$) to handle zero gains.  After getting the scores $\{\text{score}(j):j=1,\ldots, t\}$, the variable $x_{i_j}$ with the largest $\text{score}(j)$ is branched on.
\end{definition}

Empirically, strong branching often produces search trees with significantly smaller sizes than other branching heuristics do, as reviewed in~\citep{ACHTERBERG200542}. Nevertheless, since strong branching requires solving two LP problems for each candidate variable, its per-node computational costs are in general unacceptable. Hence, state-of-the-art MILP solvers such as SCIP~\citep{a2e5d9e8f52a47c183e8ad12e65499e5} usually adopt so-called \emph{hybrid branching} heuristics\footnote{In SCIP, the default branching rule is reliability branching proposed in \citep{ACHTERBERG200542}, which is an instance of hybrid branching.}, i.e., heuristics that combine strong branching and some computationally cheaper branching strategies.

\section{Theoretical results}

\subsection{Representing MILP problems as graphs}

Following~\citep{chen2025rethinkingcapacitygraphneural,chen2023representingmixedintegerlinearprograms}, we may use a \textbf{bipartite graph} to represent an arbitrary MILP problem uniquely. Given an MILP problem of form \eqref{eq:def-milp}, its bipartite graph representation is defined as following.

\begin{definition}[Bipartite graph representation of MILP, following \citep{chen2025rethinkingcapacitygraphneural,chen2023representingmixedintegerlinearprograms}]\label{def:bipartite-rep}
    The \textbf{bipartite graph representation} of MILP problem \eqref{eq:def-milp} is defined as $G=(V,W,\bm{A},F_V,F_W)$. 
    \begin{itemize}
        \item $V=\{1,\ldots, m\}$ is called the set of \textbf{constraint nodes} of $G$. A node $i\in V$ corresponds to an inequality constraint $\sum_{j=1}^n A_{ij}x_j\leqslant b_i$.
        \item $W=\{1,\ldots, n\}$ is called the set of \textbf{variable nodes} of $G$. A node $j\in W$ corresponds to a variable $x_j$.
        \item The edge set of $G$ is defined by the coefficient matrix $\bm{A}$ as in \eqref{eq:def-milp}. For $i\in V$ and $j\in W$, $(i,j)$ is an edge of $G$ if and only if $A_{ij}\ne 0$, and in this case $A_{ij}$ is the edge weight corresponding to $(i,j)$. Suppose there are $E$ many non-zero entries in $\bm{A}$.
        \item $F_V$ represents constraint node features. For $i\in V$, its feature is defined as $F_i=b_i$.
        \item $F_W$ represents variable node features. For $j\in W$, its feature is defined as $F_j=(c_j,l_j,u_j,1_{j\in\{1,\ldots,r\}})$, where $1_{j \in \mathcal{I}}$ is an indicator function.
    \end{itemize}
\end{definition}

\subsection{Subgraph neural networks for bipartite graphs}

To bridge the gap between efficient MPNNs and expressive but expensive 2-FGNNs, we focus on \textbf{Node-Anchored Subgraph GNNs}. Later, we simply call it Subgraph GNNs. This architecture enhances expressivity by running a standard GNN on subgraphs conditioned on a central ``anchor'' node. In the context of branching, the anchor node corresponds to the candidate variable being evaluated.

A great number of works (such as ~\citep{li2020distance,zhang2021nested,you2021identityawaregraphneuralnetworks,zhang2023completeexpressivenesshierarchysubgraph,bevilacqua2022equivariant}) have formally defined and studied subgraph GNNs on general graphs. Since our main purpose is to deploy the architecture of subgraph GNNs on the task of solving MILPs, we are motivated to adapt the existing definition of subgraph GNNs to the special case of bipartite graphs. For our purpose, it suffices to consider bipartite graph representations of MILPs as defined in Definition \ref{def:bipartite-rep}.

\begin{definition}[Subgraph GNNs for bipartite graphs]\label{def:subgraph-gnn}
    Let $G=(V,W,\bm{A},F_V,F_W)$ be the bipartite graph representation of an MILP problem. A subgraph GNN performs the following operations on $G$.
    \begin{itemize}
        \item For each variable node $j'\in W$ as an anchor, initialize the feature vectors of each $i\in V$ and $j\in W$ as 
        \begin{align}
            s^0_{ij'}&=p^0(F_i,F_{j'}),\label{eq:init0-sub}\\
            t^0_{jj'}&=q^0(F_j,F_{j'},1_{j=j'}).\label{eq:init1-sub}
        \end{align}
        \item For each layer $l=1,\ldots, L$, update $s^{l-1}_{ij'}$ and $t^{l-1}_{jj'}$ for each $i\in V$ and $j,j'\in W$ to
        \begin{align}
            s_{ij'}^{l}&=p^l\left(s_{ij'}^{l-1},\sum_{j\in \mathcal{N}_W(i)}f^l(t_{jj'}^{l-1},A_{ij})\right),\\
            t_{jj'}^{l}&=q^l\left(t_{jj'}^{l-1},\sum_{i\in\mathcal{N}_V(j)}g^l(s_{ij'}^{l-1},A_{ij})\right),
        \end{align}
        where $\mathcal{N}_W(i)\subseteq W$ and $\mathcal{N}_V(j)\subseteq V$ refer to sets of neighbors of constraint node $i$ and variable node $j$ respectively.
        \item Compute the final feature vectors of variable node $j'\in W$ by
        \begin{align}
            y_{j'}=r\left(\sum_{i\in V}s_{ij'}^L,\sum_{j\in W}t_{jj'}^L\right).\label{eq:final-layer}
        \end{align}
    \end{itemize}
    In the above definition, $\{p^l,q^l\}_{l=0}^L$, $\{f^l,g^l\}_{l=1}^L$ and $r$ are learnable functions usually modeled by MLPs.
\end{definition}

It is now helpful to compare the above definition of subgraph GNNs with that of message-passing GNNs (MP-GNNs) or second-order folklore GNNs (2-FGNNs).

\begin{definition}[Message-passing GNNs for bipartite graphs, following \citep{chen2025rethinkingcapacitygraphneural,chen2023representingmixedintegerlinearprograms}]\label{def:mpnn}
    Let $G=(V,W,\bm{A},F_V,F_W)$ be the bipartite graph representation of an MILP problem. A message-passing GNN performs the following operations on $G$.
    \begin{itemize}
        \item For each $i\in V$ and $j\in W$, initialize their feature vectors as
        \begin{align}
            s_i^0&=p^0(F_i),\\
            t_j^0&=q^0(F_j).
        \end{align}
        \item For each layer $l=1\ldots, L$, update $s_i^{l-1}$ and $t_j^{l-1}$ for each $i\in V$ and $j\in W$ to
        \begin{align}
            s_i^l&=p^l\left(s_i^{l-1},\sum_{j\in\mathcal{N}_W(i)} f^l(t^{l-1}_j,A_{ij})\right),\\
            t_j^l&=q^l\left(t_j^{l-1},\sum_{i\in\mathcal{N}_V(j)}g^l(s_i^{l-1},A_{ij})\right).
        \end{align}
        \item Compute the final feature vectors of variable node $j\in W$ by
        \begin{align}
            y_j=r\left(\sum_{i\in V}s_i^L,\sum_{j'\in W}t_{j'}^L, t_j^L\right).
        \end{align}
    \end{itemize}
    Here, $\{p^l,q^l\}_{l=0}^L$, $\{f^l,g^l\}_{l=1}^L$ and $r$ are also learnable functions usually modeled by MLPs.
\end{definition}

\begin{definition}[Second-order folklore GNNs for bipartite graphs, following \citep{chen2025rethinkingcapacitygraphneural}]\label{def:fwl2}
    Let $G=(V,W,\bm{A},F_V,F_W)$ be the bipartite graph representation of an MILP problem. A second-order folklore GNN performs the following operations on $G$.
    \begin{itemize}
        \item For each variable node $j'\in W$, initialize the feature vectors of each $i\in V$ and $j\in W$ as 
        \begin{align}
            s^0_{ij'}&=p^0(F_i,F_{j'},A_{ij'}),\\
            t^0_{jj'}&=q^0(F_j,F_{j'},1_{j=j'}).
        \end{align}
        \item For each layer $l=1,\ldots, L$, update $s^{l-1}_{ij'}$ and $t^{l-1}_{jj'}$ for each $i\in V$ and $j,j'\in W$ to
        \begin{align}
            s_{ij'}^{l}&=p^l\left(s_{ij'}^{l-1},\sum_{j\in W}f^l(t_{jj'}^{l-1},s_{ij}^{l-1})\right),\\
            t_{jj'}^{l}&=q^l\left(t_{jj'}^{l-1},\sum_{i\in V}g^l(s_{ij'}^{l-1},s_{ij}^{l-1})\right).
        \end{align}
        \item Compute the final feature vectors of variable node $j'\in W$ by
        \begin{align}
            y_{j'}=r\left(\sum_{i\in V}s_{ij'}^L,\sum_{j\in W}t_{jj'}^L\right).
        \end{align}
    \end{itemize}
    The notations $\{p^l,q^l\}_{l=0}^L$, $\{f^l,g^l\}_{l=1}^L$ and $r$ follow the above two definitions.
\end{definition}

Comparing Definition \ref{def:subgraph-gnn} and Definition \ref{def:mpnn}, one may see that subgraph GNNs can be understood as running an independent MP-GNN for each $j'\in V$. Those MP-GNNs only differ in their initial layers due to the different inputs $F_{j'}$ and $1_{j=j'}$ to $p^0$ and $q^0$, as shown in equations \eqref{eq:init0-sub} and \eqref{eq:init1-sub}. Consequently, while MP-GNN has a time complexity of $O(EL)$ where $E$ is the number of non-zero elements in $\bm{A}$, the time complexity of subgraph GNN is $O(nEL)$ due to the need of running $n$ independent MP-GNNs.

On the other hand, 2-FGNN and subgraph GNN have the common feature that they both maintain a feature vector for each node pair $(i,j')\in V\times W$ and $(j,j')\in W\times W$ at the initial layer. Nevertheless, the key difference between them lies in the update layers. For example, to update $s_{ij'}^{l-1}$ for $i\in V$ and $j'\in W$, subgraph GNN leverages information of $t_{jj'}^{l-1}$ where $j$ is a neighbor of $i$, along with the edge weight between $i$ and $j$; contrarily, 2-FGNN uses information from \emph{all} $t_{jj'}^{l-1}$ where $j$ is not restricted to be a neighbor of $i$. Similar things go for the update of $t_{jj'}^{l-1}$. Therefore, we may say that compared with 2-FGNN, subgraph GNN is aware of \emph{local} or \emph{neighborhood} information. As a result, the time complexity of 2-FGNN is $O(n^2(m+n)L)$, which is typically higher than the $O(nEL)$ complexity of subgraph GNN, especially when the matrix $\bm{A}$ is sparse, i.e., $E\ll mn$.

\subsection{Subgraph GNNs provably approximate strong branching scores}

Our main theoretical result is as indicated in the title of this subsection.

\begin{theorem}\label{thm:main}
Let $\mathcal{D}$ be a finite dataset where each instance is a bipartite graph representation $G=(V,W,\bm{A},F_V,F_W)$ of an MILP problem of the form \eqref{eq:def-milp}. For an instance $G\in\mathcal{D}$, let $x_{i_1},\ldots, x_{i_t}$ be the set of variables from which a branching variable must be selected. Assume every $G\in\mathcal{D}$ satisfies the condition that for each $j=1,\ldots, t$, the strong branching score $\mathrm{score}(j)$, as defined by equation \eqref{eq:sb-score}, is finite. Then, for any $\delta>0$, there exists a Subgraph GNN $\mathcal{G}_{\text{sub}}$ such that
\begin{align}
|y_{i_j}(G) - \mathrm{score}(j) |<\delta, \quad \forall \ G\in\mathcal{D} \text{\textrm{ and }} j=1,\ldots, t,
\end{align}
where $y_{i_j}(G)$ denotes the final feature vector of variable node $i_j$ produced by $\mathcal{G}_{\text{sub}}$ acting on $G$, as defined in equation \eqref{eq:final-layer}.
\end{theorem}

Before proving the above theorem, we present a simple lemma which would greatly simplify our proof.

\begin{lemma}\label{lemma:bridging}
    Define a ``modified'' subgraph GNN $\mathcal{G}_{\text{sub}}'$ with layer parameter $(L_1,L_2)$ as follows. In equation \eqref{eq:init1-sub} of Definition \ref{def:subgraph-gnn}, instead of using $F_{j'}$ as an input to $q^0$, replace it by the final feature vector $\tilde{y}_{j'}$ of $j'$ produced by an MP-GNN $\tilde{F}$ with $L_1$ update layers, for each $j'\in W$. The update and final layers of $F'$ exactly follow those in Definition \ref{def:subgraph-gnn}, and the total number of update layers is $L_2$. Then, there exists a subgraph GNN $\mathcal{G}_{\text{sub}}$ following Definition \ref{def:subgraph-gnn}, such that $\mathcal{G}_{\text{sub}}$ and $F'$ are the same function.
\end{lemma}

The proof to Lemma \ref{lemma:bridging} is simple. Actually, we may explicitly construct such $\mathcal{G}_{\text{sub}}$. To begin, we first slightly modify the update layers of a subgraph GNN to enable ``global aggregation'',
\begin{align}
    s^l_{ij'}&=p^l\left(s_{ij'}^{l-1},\sum_{j\in \mathcal{N}_W(i)} f^l(t^{l-1}_{jj'},A_{ij}), \sum_{j\in W} t_{jj'}^{l-1},\sum_{i\in V} s_{ij'}^{l-1}\right),\\
    t^l_{jj'}&=q^l\left(t_{jj'}^{l-1},\sum_{i\in \mathcal{N}_V(j)} g^l(s^{l-1}_{ij'},A_{ij}),\sum_{j\in W} t_{jj'}^{l-1},\sum_{i\in V} s_{ij'}^{l-1}\right).
\end{align}
It is well known that introducing global aggregation does not increase the expressive power of subgraph GNNs (see, for example, Theorem 4.4 of \citep{zhang2023completeexpressivenesshierarchysubgraph}). Further, computing those global aggregation terms requires $O(n(m+n)L)$ time, which does not increase the asymptotic time complexity $O(nEL)$ of original subgraph GNNs. With global aggregation enabled, we can easily use a subgraph GNN $\mathcal{G}_\text{sub}$ to simulate the ``modified'' subgraph GNN in Lemma \ref{lemma:bridging}. Let the initial layer of $\mathcal{G}_\text{sub}$ be 
\begin{align}
    s_{ij'}^0&=(F_i,\tilde{p}^0(F_i),F_{j'}),\label{eq:prooflemma1}\\
    t_{jj'}^0&=(F_j,\tilde{q}^0(F_j),F_{j'},1_{j=j'}),\label{eq:prooflemma2}
\end{align}
where $\tilde{p}^0$ and $\tilde{q}^0$ are functions appearing in the initial layer of MP-GNN $\mathcal{G}_{mp}$. The first $(L_1+1)$ update layers of $\mathcal{G}_\text{sub}$ simply simulate $\mathcal{G}_{mp}$ in $\tilde{p}^0(F_i)$ and $\tilde{q}^0(F_j)$, while copying other information in $s_{ij'}^0$ and $t_{jj'}^0$, for $i\in V$ and $j,j'\in W$. After those layers, we have
\begin{align}
    t_{jj'}^{L_1+1}=(F_j, \tilde{y}_j, F_{j'}, 1_{j=j'}).
\end{align}
In the next layer, we may let
\begin{align}
    t_{jj'}^{L_1+2}=\left(F_j,\sum_{j\in W} \tilde{y}_j\cdot 1_{j=j'},1_{j=j'}\right)=(F_j,\tilde{y}_{j'},1_{j=j'}).
\end{align}
In other words, the first $(L_1+2)$ update layers of $\mathcal{G}_{\text{sub}}$ recover the input to the update layers of the ``modified'' subgraph GNN $\mathcal{G}_{\text{sub}}'$ in Lemma~\ref{lemma:bridging}. After that, $\mathcal{G}_{\text{sub}}$ uses another $L_2$ update layers and a final layer to simulate $\mathcal{G}_{\text{sub}}'$. It is straightforward that $\mathcal{G}_{\text{sub}}$ and $\mathcal{G}_{\text{sub}}'$ describe the same function.

We are now ready to prove Theorem \ref{thm:main}. Our proof relies on two key results in \citep{chen2023representinglinearprogramsgraph}, which we state below.

\begin{theorem}\label{thm:obj}
    Let $\mathcal{D}$ be a finite dataset consisting of bipartite graph representations of linear programs (which are simply bipartite graph representations of MILPs with no integral constraints). The LP corresponding to any $G\in\mathcal{D}$ has a finite optimal objective value. Then for any $\delta>0$, there exists an MP-GNN, such that
    \begin{align}
        \left|\sum_{j\in W} y_j(G)-\Phi_\text{obj}(G)\right|<\delta,\quad \forall \ G\in\mathcal{D}.
    \end{align}
    Here, $y_j$ is the final representation of node $j$, and $\Phi_\text{obj}(G)$ means the optimal objective value of the LP corresponding to $G$.
\end{theorem}

\begin{theorem}\label{thm:sol}
    Let $\mathcal{D}$ be a measurable set of bipartite graph representations of LPs. For any $\epsilon,\delta>0$, there exists an MP-GNN, such that the measure of
    \begin{align}
        \{G\in\mathcal{D}: \|y(G)-\Phi_{\text{sol}}(G)\|>\delta\}
    \end{align}
    is smaller than $\epsilon$. Here $\Phi_\text{sol}(G)\in\mathbb{R}^n$ is the vector of optimal solution to the LP corresponding to $G$, and $y(G)=(y_1(G),\ldots, y_n(G))$. Especially, if $\mathcal{D}$ is a finite dataset, then for any $\delta>0$, there exists an MP-GNN, such that
    \begin{align}
        \|y(G)-\Phi_{\text{sol}}(G)\|<\delta,\quad\forall \ G\in\mathcal{D}.
    \end{align}
\end{theorem}

The rest of this subsection is devoted  to proving Theorem \ref{thm:main}.

\begin{proof}
    By Lemma \ref{lemma:bridging}, we only need to find a ``modified'' subgraph GNN $\mathcal{G}_{\text{sub}}'$ for any $\delta>0$, such that
    \begin{align}
        |y_{i_j}'(G)-\mathrm{score}(j)|<\delta,\quad\forall\ G\in\mathcal{D}\text{ and }j=1,\ldots, t.
    \end{align}
    This is easy to achieve. Actually, by Theorems \ref{thm:obj} and \ref{thm:sol}, for any $\delta_1,\delta_2>0$, we may find two MP-GNNs $\mathcal{G}_{\text{mp},1}$ and $\mathcal{G}_{\text{mp},2}$, such that
    \begin{gather}
        \left|\sum_{k\in W} y_k^{\mathcal{G}_{\text{mp},1}^0}(G)-\bm{c}^T\bm{x}^*\right|<\delta_1,\quad \forall \ G\in\mathcal{D},\\
        \|y^{\mathcal{G}_{\text{mp},2}^0}(G)-\bm{x}^*\|<\delta_2,\quad \forall \ G\in\mathcal{D}.
    \end{gather}
    Here $\bm{x}^*$ means the optimal solution to the LP relaxation of MILP \eqref{eq:def-milp}, as before, and . For any $\mathcal{D}$ with a finite size, we can always choose a proper $\delta_2>0$, such that for any $G\in\mathcal{D}$ and any non-integral $x^*_k$ ($k\in\{1,\ldots, n\}$), we have
    \begin{align}
        \big\lfloor y^{\mathcal{G}_{\text{mp},2}^0}_k(G)\big\rfloor=\lfloor x^*_k\rfloor,\quad \big\lceil y^{\mathcal{G}_{\text{mp},2}^0}_k(G)\big\rceil=\lceil x^*_k\rceil.
    \end{align}
    We may then let the input to the update layers of ``modified'' subgraph GNN $\mathcal{G}_{\text{sub}}'$ be
    \begin{align}
        s_{ij'}^0&=p^0(b_i),\\
        t_{kj'}^0&=q^0\left(c_k,l_k,u_k,\big\lfloor y^{\mathcal{G}_{\text{mp},2}^0}_{j'}(G)\big\rfloor,\big\lceil y^{\mathcal{G}_{\text{mp},2}^0}_{j'}(G)\big\rceil,1_{k=j'}\right),
    \end{align}
    for every $i\in V$, $k\in W$ and $j'\in\{i_1,\ldots, i_t\}$. Applying Theorem \ref{thm:obj} again, we may find that for any $\delta_3,\delta_4>0$, there exists MP-GNNs $\mathcal{G}_{\text{mp},+}$ and $\mathcal{G}_{\text{mp},-}$ such that for any $G\in\mathcal{D}$ and for any candidate variable $x_{j'}$ where $j'=i_j$, $j=1,\ldots, t$, we have
    \begin{gather}
        \left|\sum_{k\in W} y_k^{\mathcal{G}_{\text{mp},+}}(G)-\bm{c}^T\bm{x}^*-\Delta_{j}^+\right|<\delta_3,\\
        \left|\sum_{k\in W} y_k^{F^-}(G)-\bm{c}^T\bm{x}^*-\Delta_{j}^-\right|<\delta_4.
    \end{gather}
    One may take $\mathcal{G}_{\text{sub}}'$ as a subgraph GNN which performs the updated layers of $\mathcal{G}_{\text{mp},+}$ and $\mathcal{G}_{\text{mp},-}$ in parallel for every $j'\in\{i_1,\ldots, i_t\}$ independently. Assume that $L$ is the larger one of the numbers of update layers of $\mathcal{G}_{\text{mp},+}$ and $\mathcal{G}_{\text{mp},-}$. We have for $\mathcal{G}_{\text{sub}}'$
    \begin{gather}
        \left|\sum_{k\in W} t^{L+}_{kj'}-\sum_{k\in W}y_k^{\mathcal{G}_{\text{mp},1}^0}(G)-\Delta_j^+\right|<\delta_1+\delta_3,\\
        \left|\sum_{k\in W} t^{L-}_{kj'}-\sum_{k\in W}y_k^{\mathcal{G}_{\text{mp},1}^0}(G)-\Delta_j^-\right|<\delta_1+\delta_4.
    \end{gather}
    where $j'=i_j$, $t^{L+}_{kj'}$ and $t^{L-}_{kj'}$ refer to the $t_{kj'}^L$ produced by $\mathcal{G}_{\text{mp},+}$ and $\mathcal{G}_{\text{mp},-}$ for each $j'$ respectively. Notice that $y_k^{\mathcal{G}_{\text{mp},1}^0}(G)$ can be copied at each step of $\mathcal{G}_{\text{sub}}'$, and it is permitted to use the information from $y_k^{\mathcal{G}_{\text{mp},1}^0}(G)$ at the final layer. The final layer of $\mathcal{G}_{\text{sub}}'$ is
    \begin{align}
        y_{j'}=f\left(\sum_{k\in W} t^{L+}_{kj'}-\sum_{k\in W}y_k^{\mathcal{G}_{\text{mp},1}^0}(G),\sum_{k\in W} t^{L-}_{kj'}-\sum_{k\in W}y_k^{\mathcal{G}_{\text{mp},1}^0}(G)\right),
    \end{align}
    for each $j'\in\{i_1,\ldots, i_t\}$. Since $f$ is continuous with respect to its arguments, by choosing sufficiently small $\delta_1,\delta_3,\delta_4>0$, we have
    \begin{align}
        \left|y_{i_j}-f(\Delta_j^+,\Delta_j^-)\right|<\delta,\quad \forall \ G\in\mathcal{D}\text{ and }j=1,\ldots, t,
    \end{align}
    or simply $|y_{i_j}-\text{score}(j)|<\delta$. This makes the proof.
\end{proof}

\section{Experiments}


As established in the theoretical analysis, (Node-Anchored) Subgraph GNNs possess the expressivity required to approximate Strong Branching. In this section, we investigate whether this theoretical advantage translates to practical efficiency in MILP solving. Specifically, we aim to quantify the trade-off between the superior decision-making accuracy of Subgraph GNNs and their increased computational overhead compared to standard MPNNs.

We closely followed the experimental pipeline from \cite{gasse2019exactcombinatorialoptimizationgraph} to train and evaluate various configurations that may benefit the branching performance of GNNs. The details of this pipeline can be found in Section 5 of \cite{gasse2019exactcombinatorialoptimizationgraph}. Here, we briefly explain the pipeline and our specific implementations.

\subsection{Setup}

\paragraph{Benchmarks and Datasets} We focus on the following four types of MILP problems (benchmarks): Set Covering (\textsc{setcover}), Maximum Independent Set (\textsc{indset}), Combinatorial Auction (\textsc{cauction}), and Capacitated Facility Location (\textsc{facilities}). Note that, as mentioned before, one motivating reason for developing learning-to-branch is to have data-specific models that can potentially perform better than general-purpose heuristics. Therefore, the training and evaluation of all models are carried out for each problem separately.


To assess scalability, we generate instances at three difficulty levels for each problem type: Small (Easy), Medium, and Large (Hard). This stratification allows us to observe how the computational cost of the GNN models scales relative to the problem size. We use the \texttt{Ecole} library \citep{prouvost2020ecole} to sample instances and generate \textbf{branching samples} (state-action pairs containing the current bipartite graph and strong branching scores) for training.

\begin{table}[htbp]
\centering
\caption{Problem Dimensions (Variables $\times$ Constraints) for Benchmark Datasets}
\label{tab:sizes}
\begin{tabular}{lccc}
\toprule
Problem Type & Small & Medium & Large \\
\midrule
Set Covering & $1000 \times 500$ & $1000 \times 1000$ & $1000 \times 2000$ \\
Comb. Auction & $500 \times 100$ & $1000 \times 200$ & $1500 \times 300$ \\
Facilities & $50 \times 50$ & $100 \times 100$ & $150 \times 150$ \\
Indep. Set & 500 nodes & 1000 nodes & 1500 nodes \\
\bottomrule
\end{tabular}
\end{table}


\paragraph{Configurations} We explore different configurations of learning-to-branch models. The first factor is choice of \textbf{base GNN models}, including vanilla MPNNs and subgraph GNNs. \cite{gasse2019exactcombinatorialoptimizationgraph} has shown empirically that a specific 2-layer MPNN can achieve comparable or better accuracy and runtime in branching. In this work, we implemented a higher-layer MPNN and a subgraph GNN. Given the theoretical results that subgraph GNNs have time complexity between that of MPNNs and 2-FGNNs, and that subgraph GNNs are expressive enough for branching, we would like to see if the experimental results of the two base models match the theory.

The second factor is the choice of \textbf{loss functions}. The branching problem can be seen as a task of different natures. It can be regarded as a \textbf{classification task}, i.e., selecting the best candidate variable. Therefore, the Cross Entropy loss (\textsc{rank}) can be adopted to provide supervision signal for this task. Branching can also be viewed as a \textbf{regression task}, where the goal is to predict the strong branching score directly. From this point of view, the Regression loss (\textsc{scores}) can be adopted. Additionally, branching is a \textbf{ranking task} -- the essential goal is to rank the variables according to their strong branching scores. In this case, the Pairwise Ranking loss (\textsc{pairwise}) can be adopted. Different loss functions may induce different \textit{inductive biases}, and the more the loss function reflects the essence of the branching problem, the better performance is expected. Intuitively, compared to the Cross Entropy loss (\textsc{rank}), the Regression loss (\textsc{scores}) provides the direct target, i.e., the strong branching score, as the supervision signal. However, this might introduce numerical issues as the branching scores of different variables could be very close. On the other hand, the Pairwise Ranking loss (\textsc{pairwise}) may facilitate more comprehensive comparison between candidate variables, hence possibly giving better guidance.

The options under the two factors are freely combined to give \(2 \cdot 3 = 6\) configurations in total. We train and evaluate the model of every configuration individually for each problem type.

\paragraph{Traning Process} Branching samples are the primary data for training. Given the information of the current branching node and historical statistics as input, the model is expected to predict the best strong branching candidate and is optimized against one of the losses mentioned above. As a lighter model, MPNNs are trained for up to 500 epochs, while subgraph GNNs are trained for up to 200 epochs. The checkpoints with the lowest validation loss are kept for evaluation.

\subsection{Evaluation Metrics}

Our evaluation focuses on two complementary dimensions: prediction quality and solving efficiency. Since in the branch-and-bound framework, the MILP is solved through searching rather than in one shot, higher accuracy usually results in a smaller search tree and hence faster convergence. However, with this correlation in mind, we are still interested in how subgraph GNNs, which are stronger but less efficient than MPNNs in theory, may compensate for the overall solving time through higher accuracy.

In this regard, there are three evaluation metrics. The first one, \textbf{\textsc{Accuracy}}, refers to the average accuracy that the model selects the candidate variable with the greatest strong branching score on unseen test branching samples. This metric is branching sample-specific, while the following two metrics are instance-specific, meaning that the trained model is integrated into the solver to evaluate its performance in solving the whole instance. The \textbf{\textsc{Node}} metric represents the node count of the search tree produced by the method, and the \textbf{\textsc{Time}} metric measures the solving time in seconds. The latter two metrics are also tested for SCIP's default branching heuristic, ``reliable pseudocost'' (\textsc{rpb}), to provide a comparison between learning to branch and hand-crafted heuristics.

\subsection{Experiment Results}

The \textsc{Accuracy} results are presented in \cref{tab:accuracy}, and the \textsc{Time} and \textsc{Node} results are presented in \cref{table:setcover-solver,table:indset-solver,table:cauction-solver,table:facilities-solver}, respectively. For \cref{table:setcover-solver,table:indset-solver,table:cauction-solver,table:facilities-solver}, in each cell, the upper line represents the results of the corresponding configuration, while the lower line represents the results of the heuristics.

To ensure a rigorous and fair comparison, we re-evaluated the \textsc{RPB} alongside each GNN model configuration. Consequently, the reported performance metrics for the heuristic vary slightly across different tables. This variation reflects real-time fluctuations in hardware load and the impact of time limits on node counting during different experimental runs, ensuring that the GNN models are compared against the heuristic under identical machine conditions.

The results are written in the form \(A \pm B \%\), meaning that the logarithm of the metric \(X\) (either \textsc{Time} or \textsc{Node}) has mean \(\log A\) and standard deviation \(B \%\).


\begin{table}[htbp]
\centering
\caption{Branching Prediction \textsc{Accuracy} on Test Sets}
\label{tab:accuracy}
\begin{tabular}{lcccc}
\toprule
Setting & \textsc{SetCover} & \textsc{IndSet} & \textsc{CAuction} & \textsc{Facilities} \\
\midrule
MPNN + \textsc{scores} & 0.586 & 0.468 & 0.570 & 0.636 \\
\,+ \textsc{rank} & 0.618 & 0.580 & 0.600 & 0.670 \\
\,+ \textsc{pairwise} & 0.643 & 0.582 & 0.610 & 0.664 \\
\midrule
Subgraph + \textsc{scores} & 0.411 & 0.353 & 0.562 & 0.577 \\
\,+ \textsc{rank} & 0.541 & 0.550 & 0.600 & 0.649 \\
\,+ \textsc{pairwise} & \textbf{0.667} & \textbf{0.611} & \textbf{0.649} & \textbf{0.671} \\
\bottomrule
\end{tabular}
\end{table}

\begin{table}[htbp]
\centering\tiny
\setlength{\tabcolsep}{3pt}
\caption{Set Covering. \textbf{GNN Model (Top)} vs. \textbf{Heuristic Baseline (Bottom)}.}
\label{table:setcover-solver}
\begin{tabular}{lcccccc}
\toprule
\multirow{2}{*}{Setting} & \multicolumn{2}{c}{Small (s)} & \multicolumn{2}{c}{Medium (m)} & \multicolumn{2}{c}{Large (l)} \\ 
\cmidrule(lr){2-3} \cmidrule(lr){4-5} \cmidrule(lr){6-7}
& \textsc{Time}& \textsc{Node}& \textsc{Time}& \textsc{Node}& \textsc{Time}& \textsc{Node}\\
\midrule
MPNN + \textsc{scores} & 
  \makecell{3.99 $\pm$ 82\% \\ \textit{4.94 $\pm$ 65\%}} & 
  \makecell{147 $\pm$ 125\% \\ \textit{49 $\pm$ 196\%}} & 
  \makecell{15.27 $\pm$ 106\% \\ \textit{14.79 $\pm$ 75\%}} & 
  \makecell{709 $\pm$ 128\% \\ \textit{519 $\pm$ 183\%}} & 
  \makecell{48.0 $\pm$ 129\% \\ \textit{39.0 $\pm$ 98\%}} & 
  \makecell{2272 $\pm$ 145\% \\ \textit{2324 $\pm$ 175\%}} \\
\midrule
\, + \textsc{rank} & 
  \makecell{4.26 $\pm$ 80\% \\ \textit{5.20 $\pm$ 65\%}} & 
  \makecell{162 $\pm$ 121\% \\ \textit{63 $\pm$ 196\%}} & 
  \makecell{16.47 $\pm$ 95\% \\ \textit{16.37 $\pm$ 68\%}} & 
  \makecell{758 $\pm$ 119\% \\ \textit{709 $\pm$ 168\%}} & 
  \makecell{58.1 $\pm$ 132\% \\ \textit{46.2 $\pm$ 102\%}} & 
  \makecell{2741 $\pm$ 146\% \\ \textit{3117 $\pm$ 166\%}} \\
\midrule
\, + \textsc{pairwise} & 
  \makecell{3.75 $\pm$ 68\% \\ \textit{4.98 $\pm$ 59\%}} & 
  \makecell{133 $\pm$ 107\% \\ \textit{52 $\pm$ 170\%}} & 
  \makecell{16.42 $\pm$ 112\% \\ \textit{16.21 $\pm$ 81\%}} & 
  \makecell{745 $\pm$ 136\% \\ \textit{591 $\pm$ 187\%}} & 
  \makecell{99.2 $\pm$ 151\% \\ \textit{51.4 $\pm$ 105\%}} & 
  \makecell{5101 $\pm$ 162\% \\ \textit{3538 $\pm$ 169\%}} \\
\midrule
Subgraph + \textsc{scores} & 
  \makecell{32.5 $\pm$ 146\% \\ \textit{5.11 $\pm$ 66\%}} & 
  \makecell{471 $\pm$ 156\% \\ \textit{58 $\pm$ 192\%}} & 
  \makecell{373.4 $\pm$ 138\% \\ \textit{15.1 $\pm$ 64\%}} & 
  \makecell{3998 $\pm$ 135\% \\ \textit{603 $\pm$ 163\%}} & 
  \makecell{1212 $\pm$ 127\% \\ \textit{36.8 $\pm$ 111\%}} & 
  \makecell{9309 $\pm$ 121\% \\ \textit{2024 $\pm$ 198\%}} \\
\midrule
\, + \textsc{rank} & 
  \makecell{13.5 $\pm$ 102\% \\ \textit{4.82 $\pm$ 59\%}} & 
  \makecell{182 $\pm$ 118\% \\ \textit{52 $\pm$ 175\%}} & 
  \makecell{90.2 $\pm$ 139\% \\ \textit{14.1 $\pm$ 73\%}} & 
  \makecell{920 $\pm$ 144\% \\ \textit{449 $\pm$ 203\%}} & 
  \makecell{556 $\pm$ 141\% \\ \textit{45.7 $\pm$ 101\%}} & 
  \makecell{4324 $\pm$ 138\% \\ \textit{3168 $\pm$ 164\%}} \\
\midrule
\, + \textsc{pairwise} & 
  \makecell{10.7 $\pm$ 100\% \\ \textit{5.06 $\pm$ 63\%}} & 
  \makecell{137 $\pm$ 112\% \\ \textit{59 $\pm$ 186\%}} & 
  \makecell{58.3 $\pm$ 118\% \\ \textit{14.8 $\pm$ 68\%}} & 
  \makecell{575 $\pm$ 122\% \\ \textit{489 $\pm$ 185\%}} & 
  \makecell{305 $\pm$ 157\% \\ \textit{48.9 $\pm$ 120\%}} & 
  \makecell{2271 $\pm$ 155\% \\ \textit{2941 $\pm$ 197\%}} \\
\bottomrule
\end{tabular}
\end{table}

\begin{table}[htbp]
\centering\tiny
\setlength{\tabcolsep}{3pt}
\caption{Maximum Independent Set.  \textbf{GNN Model (Top)} vs. \textbf{Heuristic Baseline (Bottom)}}
\label{table:indset-solver}
\begin{tabular}{lcccccc}
\toprule
\multirow{2}{*}{Setting} & \multicolumn{2}{c}{Small (s)} & \multicolumn{2}{c}{Medium (m)} & \multicolumn{2}{c}{Large (l)} \\ 
\cmidrule(lr){2-3} \cmidrule(lr){4-5} \cmidrule(lr){6-7}
& \textsc{Time} & \textsc{Node} & \textsc{Time} & \textsc{Node} & \textsc{Time} & \textsc{Node} \\
\midrule
MPNN + \textsc{scores} &
  \makecell{7.90 $\pm$ 149\% \\ \textit{5.27 $\pm$ 60\%}} &
  \makecell{190 $\pm$ 287\% \\ \textit{10.8 $\pm$ 171\%}} &
  \makecell{2217 $\pm$ 110\% \\ \textit{93.1 $\pm$ 96\%}} &
  \makecell{129121 $\pm$ 117\% \\ \textit{3924 $\pm$ 137\%}} &
  \makecell{Model OOT \\ \textit{1987 $\pm$ 86\%}} &
  \makecell{Model OOT \\ \textit{69494 $\pm$ 111\%}} \\
\midrule
\, + \textsc{rank} &
  \makecell{7.64 $\pm$ 158\% \\ \textit{4.15 $\pm$ 64\%}} &
  \makecell{211 $\pm$ 286\% \\ \textit{11.9 $\pm$ 181\%}} &
  \makecell{2328 $\pm$ 116\% \\ \textit{113 $\pm$ 102\%}} &
  \makecell{134601 $\pm$ 132\% \\ \textit{4480 $\pm$ 151\%}} &
  \makecell{Model OOT \\ \textit{1802 $\pm$ 85\%}} &
  \makecell{Model OOT \\ \textit{55200 $\pm$ 108\%}} \\
\midrule
\, + \textsc{pairwise} &
  \makecell{7.19 $\pm$ 174\% \\ \textit{3.65 $\pm$ 61\%}} &
  \makecell{212 $\pm$ 274\% \\ \textit{12.1 $\pm$ 195\%}} &
  \makecell{1652 $\pm$ 161\% \\ \textit{81.2 $\pm$ 111\%}} &
  \makecell{90706 $\pm$ 188\% \\ \textit{2855 $\pm$ 195\%}} &
  \makecell{Model OOT \\ \textit{1845 $\pm$ 89\%}} &
  \makecell{Model OOT \\ \textit{60882 $\pm$ 111\%}} \\
\midrule
Subgraph + \textsc{scores} &
  \makecell{10.09 $\pm$ 110\% \\ \textit{5.93 $\pm$ 74\%}} &
  \makecell{30.0 $\pm$ 169\% \\ \textit{13.6 $\pm$ 219\%}} &
  \makecell{Model OOM \\ \textit{—}} &
  \makecell{Model OOM \\ \textit{—}} &
  \makecell{Model OOM \\ \textit{—}} &
  \makecell{Model OOM \\ \textit{—}} \\
\midrule
\, + \textsc{rank} &
  \makecell{6.73 $\pm$ 89\% \\ \textit{4.44 $\pm$ 62\%}} &
  \makecell{33.7 $\pm$ 144\% \\ \textit{11.8 $\pm$ 191\%}} &
  \makecell{701 $\pm$ 131\% \\ \textit{130 $\pm$ 104\%}} &
  \makecell{1417 $\pm$ 141\% \\ \textit{3654 $\pm$ 158\%}} &
  \makecell{Model OOM \\ \textit{—}} &
  \makecell{Model OOM \\ \textit{—}} \\
\midrule
\, + \textsc{pairwise} &
  \makecell{10.18 $\pm$ 96\% \\ \textit{6.60 $\pm$ 71\%}} &
  \makecell{26.6 $\pm$ 151\% \\ \textit{12.0 $\pm$ 204\%}} &
  \makecell{922 $\pm$ 142\% \\ \textit{91.0 $\pm$ 98\%}} &
  \makecell{2224 $\pm$ 148\% \\ \textit{3416 $\pm$ 155\%}} &
  \makecell{Model OOM \\ \textit{—}} &
  \makecell{Model OOM \\ \textit{—}} \\
\bottomrule
\end{tabular}
\end{table}

\begin{table}[htbp]
\centering\tiny
\setlength{\tabcolsep}{3pt}
\caption{Combinatorial Auction. \textbf{GNN Model (Top)} vs. \textbf{Heuristic Baseline (Bottom)}}
\label{table:cauction-solver}
\begin{tabular}{lcccccc}
\toprule
\multirow{2}{*}{Setting} & \multicolumn{2}{c}{Small (s)} & \multicolumn{2}{c}{Medium (m)} & \multicolumn{2}{c}{Large (l)} \\
\cmidrule(lr){2-3} \cmidrule(lr){4-5} \cmidrule(lr){6-7}
& \textsc{Time} & \textsc{Node} & \textsc{Time} & \textsc{Node} & \textsc{Time} & \textsc{Node} \\
\midrule
MPNN + \textsc{scores} & 
  \makecell{1.31 $\pm$ 52\% \\ \textit{1.65 $\pm$ 51\%}} & 
  \makecell{77.4 $\pm$ 107\% \\ \textit{11.7 $\pm$ 135\%}} & 
  \makecell{22.4 $\pm$ 84\% \\ \textit{16.3 $\pm$ 52\%}} & 
  \makecell{1242 $\pm$ 84\% \\ \textit{1276 $\pm$ 119\%}} & 
  \makecell{515 $\pm$ 140\% \\ \textit{155 $\pm$ 117\%}} & 
  \makecell{26033 $\pm$ 142\% \\ \textit{15528 $\pm$ 140\%}} \\
\midrule
\, + \textsc{rank} & 
  \makecell{1.24 $\pm$ 42\% \\ \textit{1.76 $\pm$ 47\%}} & 
  \makecell{71.4 $\pm$ 102\% \\ \textit{13.2 $\pm$ 132\%}} & 
  \makecell{16.1 $\pm$ 85\% \\ \textit{19.9 $\pm$ 62\%}} & 
  \makecell{1543 $\pm$ 111\% \\ \textit{1762 $\pm$ 126\%}} & 
  \makecell{252 $\pm$ 133\% \\ \textit{163 $\pm$ 106\%}} & 
  \makecell{22742 $\pm$ 140\% \\ \textit{16100 $\pm$ 127\%}} \\
\midrule
\, + \textsc{pairwise} & 
  \makecell{1.23 $\pm$ 45\% \\ \textit{1.75 $\pm$ 50\%}} & 
  \makecell{71.1 $\pm$ 103\% \\ \textit{14.4 $\pm$ 132\%}} & 
  \makecell{12.9 $\pm$ 98\% \\ \textit{17.0 $\pm$ 74\%}} & 
  \makecell{1122 $\pm$ 137\% \\ \textit{1155 $\pm$ 175\%}} & 
  \makecell{168 $\pm$ 113\% \\ \textit{142 $\pm$ 93\%}} & 
  \makecell{14566 $\pm$ 120\% \\ \textit{13640 $\pm$ 113\%}} \\
\midrule
Subgraph + \textsc{scores} & 
  \makecell{2.44 $\pm$ 68\% \\ \textit{1.77 $\pm$ 48\%}} & 
  \makecell{78.6 $\pm$ 99\% \\ \textit{14.0 $\pm$ 125\%}} & 
  \makecell{89.1 $\pm$ 102\% \\ \textit{17.5 $\pm$ 52\%}} & 
  \makecell{1567 $\pm$ 105\% \\ \textit{1487 $\pm$ 121\%}} & 
  \makecell{1509 $\pm$ 102\% \\ \textit{194 $\pm$ 107\%}} & 
  \makecell{17353 $\pm$ 104\% \\ \textit{20484 $\pm$ 130\%}} \\
\midrule
\, + \textsc{rank} & 
  \makecell{2.07 $\pm$ 69\% \\ \textit{1.69 $\pm$ 53\%}} & 
  \makecell{65.6 $\pm$ 115\% \\ \textit{13.1 $\pm$ 153\%}} & 
  \makecell{76.3 $\pm$ 119\% \\ \textit{16.3 $\pm$ 62\%}} & 
  \makecell{1100 $\pm$ 121\% \\ \textit{1209 $\pm$ 147\%}} & 
  \makecell{1342 $\pm$ 94\% \\ \textit{150 $\pm$ 96\%}} & 
  \makecell{11851 $\pm$ 95\% \\ \textit{14945 $\pm$ 115\%}} \\
\midrule
\, + \textsc{pairwise} & 
  \makecell{2.16 $\pm$ 62\% \\ \textit{1.70 $\pm$ 44\%}} & 
  \makecell{65.1 $\pm$ 95\% \\ \textit{11.0 $\pm$ 129\%}} & 
  \makecell{59.9 $\pm$ 103\% \\ \textit{15.6 $\pm$ 53\%}} & 
  \makecell{966 $\pm$ 108\% \\ \textit{1100 $\pm$ 141\%}} & 
  \makecell{1182 $\pm$ 107\% \\ \textit{202 $\pm$ 124\%}} & 
  \makecell{12237 $\pm$ 104\% \\ \textit{20694 $\pm$ 144\%}} \\
\bottomrule
\end{tabular}
\end{table}

\begin{table}[htbp]
\centering\tiny
\setlength{\tabcolsep}{3pt}
\caption{Capacitated Facility Location. \textbf{GNN Model (Top)} vs. \textbf{Heuristic Baseline (Bottom)}}
\label{table:facilities-solver}
\begin{tabular}{lcccccc}
\toprule
\multirow{2}{*}{Setting} & \multicolumn{2}{c}{Small (s)} & \multicolumn{2}{c}{Medium (m)} & \multicolumn{2}{c}{Large (l)} \\
\cmidrule(lr){2-3} \cmidrule(lr){4-5} \cmidrule(lr){6-7}
& \textsc{Time} & \textsc{Node} & \textsc{Time} & \textsc{Node} & \textsc{Time} & \textsc{Node} \\
\midrule
MPNN + \textsc{scores} & 
  \makecell{27.2 $\pm$ 92\% \\ \textit{30.3 $\pm$ 88\%}} & 
  \makecell{210 $\pm$ 148\% \\ \textit{55.8 $\pm$ 229\%}} & 
  \makecell{121 $\pm$ 97\% \\ \textit{136 $\pm$ 86\%}} & 
  \makecell{389 $\pm$ 129\% \\ \textit{150 $\pm$ 176\%}} & 
  \makecell{440 $\pm$ 81\% \\ \textit{444 $\pm$ 70\%}} & 
  \makecell{339 $\pm$ 119\% \\ \textit{129 $\pm$ 145\%}} \\
\midrule
\, + \textsc{rank} & 
  \makecell{21.0 $\pm$ 84\% \\ \textit{24.8 $\pm$ 84\%}} & 
  \makecell{148 $\pm$ 140\% \\ \textit{37.1 $\pm$ 226\%}} & 
  \makecell{115 $\pm$ 91\% \\ \textit{136 $\pm$ 83\%}} & 
  \makecell{362 $\pm$ 122\% \\ \textit{142 $\pm$ 177\%}} & 
  \makecell{438 $\pm$ 86\% \\ \textit{459 $\pm$ 76\%}} & 
  \makecell{305 $\pm$ 117\% \\ \textit{113 $\pm$ 155\%}} \\
\midrule
\, + \textsc{pairwise} & 
  \makecell{25.3 $\pm$ 95\% \\ \textit{25.6 $\pm$ 93\%}} & 
  \makecell{191 $\pm$ 138\% \\ \textit{59.2 $\pm$ 208\%}} & 
  \makecell{125 $\pm$ 101\% \\ \textit{148 $\pm$ 93\%}} & 
  \makecell{377 $\pm$ 147\% \\ \textit{158 $\pm$ 195\%}} & 
  \makecell{484 $\pm$ 79\% \\ \textit{542 $\pm$ 72\%}} & 
  \makecell{376 $\pm$ 104\% \\ \textit{162 $\pm$ 138\%}} \\
\midrule
Subgraph + \textsc{scores} & 
  \makecell{28.1 $\pm$ 88\% \\ \textit{26.4 $\pm$ 79\%}} & 
  \makecell{223 $\pm$ 129\% \\ \textit{52.7 $\pm$ 209\%}} & 
  \makecell{157 $\pm$ 107\% \\ \textit{144 $\pm$ 94\%}} & 
  \makecell{492 $\pm$ 132\% \\ \textit{157 $\pm$ 186\%}} & 
  \makecell{552 $\pm$ 90\% \\ \textit{468 $\pm$ 72\%}} & 
  \makecell{420 $\pm$ 120\% \\ \textit{130 $\pm$ 145\%}} \\
\midrule
\, + \textsc{rank} & 
  \makecell{24.7 $\pm$ 89\% \\ \textit{25.9 $\pm$ 89\%}} & 
  \makecell{183 $\pm$ 134\% \\ \textit{42.0 $\pm$ 217\%}} & 
  \makecell{129 $\pm$ 96\% \\ \textit{126 $\pm$ 86\%}} & 
  \makecell{392 $\pm$ 124\% \\ \textit{117 $\pm$ 180\%}} & 
  \makecell{629 $\pm$ 85\% \\ \textit{524 $\pm$ 69\%}} & 
  \makecell{445 $\pm$ 118\% \\ \textit{150 $\pm$ 144\%}} \\
\midrule
\, + \textsc{pairwise} & 
  \makecell{29.1 $\pm$ 100\% \\ \textit{30.9 $\pm$ 93\%}} & 
  \makecell{233 $\pm$ 137\% \\ \textit{69.6 $\pm$ 206\%}} & 
  \makecell{146 $\pm$ 90\% \\ \textit{151 $\pm$ 79\%}} & 
  \makecell{501 $\pm$ 119\% \\ \textit{202 $\pm$ 166\%}} & 
  \makecell{534 $\pm$ 82\% \\ \textit{484 $\pm$ 66\%}} & 
  \makecell{400 $\pm$ 101\% \\ \textit{135 $\pm$ 127\%}} \\
\bottomrule
\end{tabular}
\end{table}

\subsection{Result Analysis}

\paragraph{Clarification}
As mentioned before, the heuristic was evaluated in all runs to provide an indicator of the machine's status. In some scenarios, for example, the evaluation process of Facilities, the results of the heuristic indicate that the machine's status across different runs varied and is not suitable for comparison. Therefore, in the following ablation study, we focus on the overall patterns shown in the results and ignore some unstable cases.

\subsubsection{Impact of GNN Architecture: Expressivity vs. Scalability}
We observe a dichotomy between prediction accuracy and solving efficiency. As shown in table \ref{tab:accuracy}, Subgraph GNNs consistently outperform MPNNs in \textsc{Accuracy} when trained with the \textsc{Pairwise} loss, particularly on \textsc{SetCover} ($0.667$ vs. $0.643$) and \textsc{IndSet} ($0.611$ vs. $0.582$). This confirms our theoretical finding that Subgraph GNNs possess superior expressive power for approximating Strong Branching.

However, this expressivity comes at a steep computational cost. In solving tasks, Subgraph GNNs generally incur significantly higher \textsc{Time} than MPNNs across all datasets (e.g., in \textsc{SetCover} Large, $305s$ vs $99s$ for the Pairwise setting). Crucially, in \textsc{IndSet} (Table \ref{table:indset-solver}), Subgraph GNNs encounter \textbf{Out-Of-Memory (OOM)} errors on Medium and Large instances. This failure mode is a direct consequence of the $O(n)$ factor overhead in memory complexity, which becomes prohibitive for denser graphs like those in \textsc{IndSet}.

Furthermore, regarding the search tree size (\textsc{Node}), Subgraph GNNs do not consistently outperform MPNNs. While they achieve smaller trees in specific hard instances (e.g., \textsc{CAuction} Large, table \ref{table:cauction-solver}), they often produce larger trees in others (e.g., \textsc{Facilities}). This suggests that while Subgraph GNNs fit the training distribution (strong branching scores) better, they may suffer from overfitting, whereas simpler MPNNs learn more robust, albeit less expressive, heuristics.

\subsubsection{Impact of Loss Functions}
The choice of loss function significantly influences performance. 
\begin{itemize}
    \item \textbf{\textsc{Scores} (Regression):} This loss consistently yields the poorest performance in both \textsc{Accuracy} and solving metrics. Direct regression of strong branching scores is numerically unstable due to the small variance in score magnitudes.
    \item \textbf{\textsc{Pairwise} vs. \textsc{Rank}:} The \textsc{Pairwise} ranking loss generally outperforms classification (\textsc{Rank}) and regression losses. By focusing on the relative ordering of candidates rather than absolute values, it provides a robust supervision signal that aligns well with the decision-making goal of branching.
\end{itemize}

\subsubsection{On Learning-to-Branch and Heuristics}

\paragraph{GNNs versus Heuristics}

The \textsc{Node} metric (i.e., the search tree size) of GNN models is generally larger than that of the heuristics. This trend is more obvious for small or medium problems, but a few settings of GNN models outperformed the heuristics in large problems, for example ``Subgraph + \textsc{pairwise}'' in Set Cover and Combinatorial Auction.

Similarly, the \textsc{Time} metric of GNN models is generally longer than that of the heuristics. There are only a few settings in which GNN models were slightly better, for example, ``MPNN + \textsc{scores}'' in Combinatorial Auction. As a consequence of more intensive CPU (for processing the instances into graphs, etc.) and GPU computation (for running GNN models), this is an expected outcome.

\paragraph{Different Datasets} From the above ablation study on other aspects, we have already seen different patterns in different types of problems. The same phenomena also appear in the comparison between learning-to-branch models and heuristics. For ``easier'' problems like Set Covering and Combinatorial Auction (see table~\ref{table:setcover-solver} and table~\ref{table:cauction-solver}), where GNN models take shorter time to train and evaluate, they often demonstrate no advantage over heuristics. For ``harder'' problems like Maximum Independent Set and Capacitated Facility Location (see table~\ref{table:indset-solver} and table~\ref{table:facilities-solver}), where both neural models and heuristics take longer time to solve, there are chances GNN models take shorter time, e.g., in medium and large problems of \textsc{facilities}. These observations indicate that GNN models may have the potential to specialize in certain types of MILP problems, especially the hard ones.

\section{Conclusion} This work establishes Node-Anchored Subgraph GNNs as a theoretically sound ``middle ground'' architecture, proving they possess sufficient expressivity to approximate Strong Branching without the cubic complexity of higher-order models. However, our empirical analysis reveals a stark reality: the $O(n)$ memory and runtime overhead of subgraph approaches remains prohibitive for many real-world MILP instances, often resulting in Out-Of-Memory errors and slower solving times compared to simpler MPNNs. These findings suggest that while high expressivity is theoretically desirable, it is not currently the primary bottleneck; future research may prioritize computational efficiency, perhaps through subgraph sampling to make learning-based branching a viable competitor to highly optimized heuristics.

\bibliographystyle{plainnat} 
\bibliography{main} 

\end{document}